\documentclass[letterpaper, 10 pt, conference]{iros}
\IEEEoverridecommandlockouts
\usepackage{graphicx}
\usepackage{booktabs}
\usepackage{amsmath}
\usepackage{amssymb}
\usepackage{multirow}
\usepackage{xcolor}
\usepackage{tabularx}
\usepackage{colortbl}  %
\usepackage{array}
\usepackage{siunitx} %
\usepackage{tcolorbox}  %
\usepackage{listings}
\usepackage{mathrsfs}
\usepackage{wrapfig}
\usepackage{upgreek}
\usepackage{makecell}
\usepackage{vcell}

\usepackage{mwe} %
\usepackage{calc} %
\usepackage{pifont} %

\usepackage{marvosym}   %
\usepackage{hyperref}

\newcommand{\yes}{\color{blue}{\ding{51}}}
\newcommand{\no}{\color{red}{\ding{55}}}
\definecolor{Gray}{gray}{0.85}
\newcommand{\method}{\mbox{{ManiGaussian++}}\xspace}

\definecolor{best}{rgb}{0.96, 0.57, 0.58}
\definecolor{second}{rgb}{0.98, 0.78, 0.57}
\definecolor{third}{rgb}{1.0, 1.0, 0.56}

\newcommand{\ourcolor}{gray!9}

\definecolor{cvprblue}{rgb}{0.21,0.49,0.74}
\usepackage{cleveref}
\title{ManiGaussian++: General Robotic Bimanual Manipulation with Hierarchical Gaussian World Model}

\author{Tengbo Yu$^{*,1}$, Guanxing Lu$^{*,1}$, Zaijia Yang$^{*,2}$, Haoyuan Deng$^{3}$, Season Si Chen$^{1}$, \\ Jiwen Lu$^{4}$, Wenbo Ding$^{1}$, Guoqiang Hu$^{3}$, Yansong Tang$^{\dagger, 1}$, Ziwei Wang$^{3}$% <-this % stops a space
\thanks{*: Equal Contribution}% <-this % stops a space
\thanks{$\dagger$: Corresponding Author}% <-this % stops a space
\thanks{$^{1}$ Tsinghua Shenzhen International Graduate School, Tsinghua University, Emails: 
        {\tt\small \{ytb23@mails.,lgx23@mails.,season.chen@,
        tang.yansong@, ding.wenbo@\}sz.tsinghua.edu.cn}}%
\thanks{$^{2}$ School of Computer Science and Technology, Hainan University, Emails: 
        {\tt\small 20213002625@hainanu.edu.cn}}%
\thanks{${^3}$ School of Electrical and Electronic Engineering, Nanyang Technological University, 
        Emails: {\tt\small \{E230112@e., gqhu@, ziwei.wang@\}
        ntu.edu.sg}}
\thanks{${^4}$ Department of Automation, Tsinghua University, Emails: 
        {\tt\small lujiwen@tsinghua.edu.cn}
}
}
\begin{document}
\maketitle
% Take a deep breath, I want you to act as an experienced academician. I'm writing a scientific paper for CVPR conference. Here is one part of my draft, please revise it to make it more fluent and compatible with CVPR conference. Only use simple words and expressions. This is very important to my career. """ """

\begin{abstract}

Multi-task robotic bimanual manipulation is becoming increasingly popular as it enables sophisticated tasks that require diverse dual-arm collaboration patterns. Compared to unimanual manipulation, bimanual tasks pose challenges to understanding the multi-body spatiotemporal dynamics. An existing method ManiGaussian~\cite{lu2025manigaussian} pioneers encoding the spatiotemporal dynamics into the visual representation via Gaussian world model for single-arm settings, which ignores the interaction of multiple embodiments for dual-arm systems with significant performance drop. In this paper, we propose \method, an extension of ManiGaussian framework that improves multi-task bimanual manipulation by digesting multi-body scene dynamics through a hierarchical Gaussian world model. To be specific, we first generate task-oriented Gaussian Splatting from intermediate visual features, which aims to differentiate acting and stabilizing arms for multi-body spatiotemporal dynamics modeling. We then build a hierarchical Gaussian world model with the leader-follower architecture, where the multi-body spatiotemporal dynamics is mined for intermediate visual representation via future scene prediction. The leader predicts Gaussian Splatting deformation caused by motions of the stabilizing arm, through which the follower generates the physical consequences resulted from the movement of the acting arm. As a result, our method significantly outperforms the current state-of-the-art bimanual manipulation techniques by an improvement of $20.2$\% in $10$ simulated tasks, and achieves $60$\% success rate on average in $9$ challenging real-world tasks. Our code is available at \url{https://github.com/April-Yz/ManiGaussian_Bimanual}.

\end{abstract}

\section{Introduction}
\label{sec:intro}

General robotic bimanual manipulation agent is trending for their immeasurable potential in completing diverse complex tasks across houses \cite{tony2023aloha,zhang2024empoweringembodiedmanipulationbimanualmobile}, hospitals~\cite{10149474}, and factories \cite{buhl2019dual}.
A bimanual system is more than a naive combination of separate single-arm agents, as it enables challenging task that entails specific collaboration patterns including simultaneously manipulating and stabilizing target objects~\cite{liu2024voxactbvoxelbasedactingstabilizing, grannen2023stabilizeactlearningcoordinate}.
Additionally, bimanual systems are often outperform unimanual agents when multiple action steps can be performed in parallel by different arms~\cite{grotz2024peract2benchmarkinglearningrobotic}. 
As a result, general bimanual manipulation system that can interact with diverse objects and environments across tasks is highly-desired in recent years. 
% Reinforcement learning?
%% 双臂操纵难度大，数据缺失
% However, teleoprating a bimanual system is extremely expensive due to the high-dimensional action space~\citep{tony2023aloha, fu2024mobilealohalearningbimanual, ding2024bunnyvisionprorealtimebimanualdexterous, cheng2024opentelevisionteleoperationimmersiveactive, wang2024dexcapscalableportablemocap, wu2024gellogenerallowcostintuitive}, and the limited demonstrations severely constrains the generalization of current bimanual systems. 
%% 说明视觉表征的重要性，是泛化性的瓶颈
Typically, a robot manipulation agent~\cite{shridhar2023peract, grotz2024peract2benchmarkinglearningrobotic} comprises a perception module that encodes the visual clues into latent representation, and a policy head that maps these representation to the robotic action space. 
% While concurrent work focuses on sharpening the policy head to improve the
However, conventional visual representations~\cite{grotz2024peract2benchmarkinglearningrobotic, liu2024voxactbvoxelbasedactingstabilizing} usually suffer from insufficient generalization ability to multi-task bimanual manipulation scenarios, which is required to mine the unstructured scene geometry across diverse tasks and objects.
% action-conditioned dynamics termed

% \begin{figure}[t]
%     \centering
%     \includegraphics[width=0.48\textwidth]{pic/teaser.pdf}
%     % \fbox{
%     %     \begin{minipage}[c][0.25\textheight][c]{0.42\textwidth}
%     %       \centering{Dummy figure}
%     %     \end{minipage}
%     %   }
%     \caption{ \textbf{Real-world Tasks.} The proposed \method is able to complete $9$ challenging real-word tasks simultaneously by mining the multi-body spatiotemporal dynamics for bimanual manipulation.
%     }
%     \vspace{-0.2cm}
%     \label{fig:teaser}
% \end{figure}

\begin{figure}[t]
    \centering
    \includegraphics[width=0.48\textwidth]{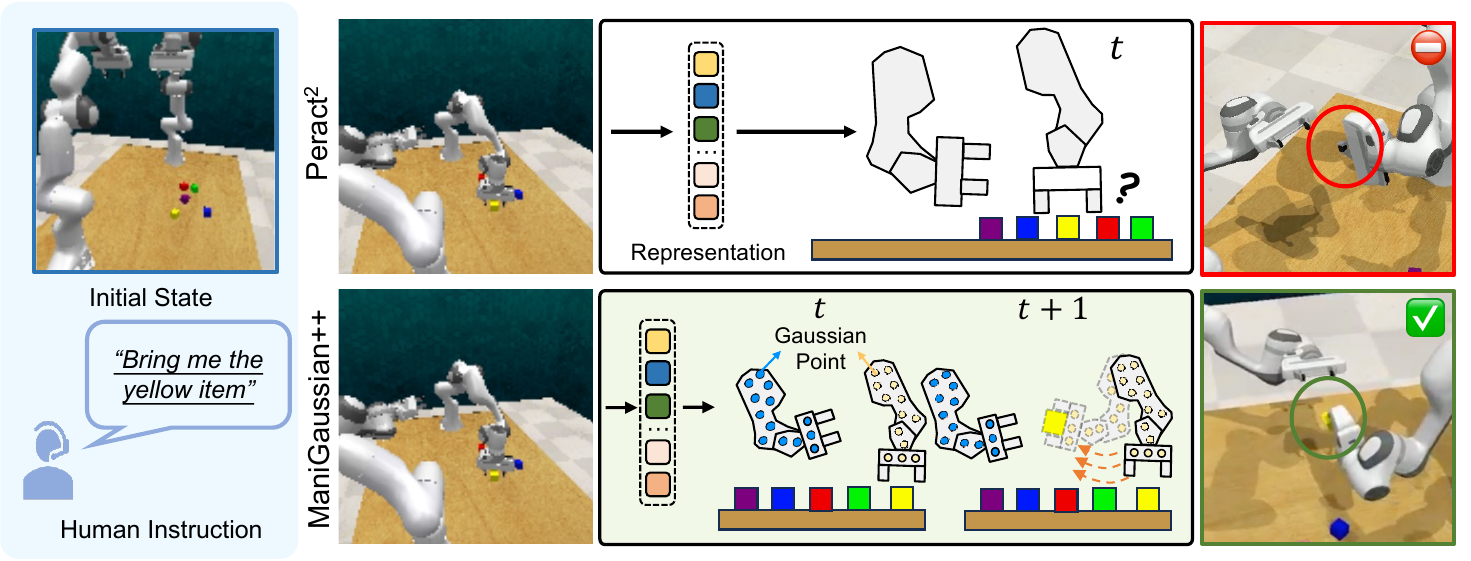}
    % \fbox{
    %     \begin{minipage}[c][0.25\textheight][c]{0.42\textwidth}
    %       \centering{Dummy figure}
    %     \end{minipage}
    %   }
    \caption{ Consider the human instruction \textit{"Bring me the yellow item"}, where the task is considered successful if the right arm handover the yellow block to the left arm. The previous method (Peract$^2$ \cite{grotz2024peract2benchmarkinglearningrobotic}) attempts to pick up the yellow block but fails to do so, while our \method completes the task successfully by explicitly encoding the scene dynamics via future scene reconstruction in Gaussian embedding space.
    }
    \vspace{-0.6cm}
    \label{fig:teaser}
\end{figure}

%% 现在大家在双臂视觉表征的进展？
To address this, existing works \cite{lu2025manigaussian, zhu2024spa, ze2023gnfactor} propose to leverage self-supervised learning to enhance the generalization ability of the visual representation for robot learning, 
% leveraging pretrained foundation models or self-supervised learning.
% 2D自监督训练视觉表征
Earlier works~\cite{parisi2022pvr,nair2022r3m,radosavovic2023realmvp} first propose to leverage pretrained 2D visual representation, which have shown initial success of visual representation learning but are constrained to relatively simple tasks due to the lack of geometric understanding like occlusion.
% 大家开始注意到3D，于是开始用NERF
To apply to more complex manipulation tasks that require 3D scene understanding, preceding methods~\cite{ze2023rl3d,zhu2024spa} attempt to model the workspace via 3D reconstruction methods like Neural Radiance Fields (NeRFs)~\cite{mildenhall2021nerf} or Gaussian Splatting~\cite{kerbl20233d}.
% 随着GS的兴起，开始用GS (ManiGaussian)
For example, ManiGaussian~\cite{lu2025manigaussian} pioneers to explicitly encode the scene dynamics via future scene reconstruction in Gaussian embedding space, which shows impressive performance in single-arm setting~\cite{james2020rlbench}.
% 但是双臂多体动力学更难学
However, learning multi-body spatiotemporal dynamics in the dual-arm system poses challenges for existing methods, thereby leads to severe performance drops in bimanual manipulation scenarios. 

In this paper, we propose a general bimanual manipulation agent named \method, which leverages hierarchical Gaussian world model to encode the multi-body spatiotemporal dynamics in dual-arm systems. 
Different from the prior ManiGaussian which only encodes coarse scene-level spatiotemporal dynamics, \method concentrates on the multi-body spatiotemporal dynamics of the dual-arm system for bimanual manipulation tasks.
Therefore, the complex interactions between the two manipulators and targets are considered to accomplish diverse collaboration patterns.
% Different from traditional arts that 
More specifically, \method first generates a task-oriented Gaussian radiance field from intermediate visual representations supervised by pre-trained vision-language models (VLMs), allowing us to assign distinct roles to the robot arms including stabilizing and acting arms for multi-body spatiotemporal dynamics modeling.
Subsequently, we mine the multi-body spatiotemporal dynamics for intermediate visual representations via future scene prediction, where a hierarchical Gaussian world model with the leader-follower architecture is utilized.
The leader predicts Gaussian Splatting deformation caused by motions of the stabilizing arm, through which the follower generates the physical consequences resulted from the movement of the acting arm.
% where the leader model captures the prestress introduced by the stabilizer, while the follower derives the final future prediction influenced by the actor conditioned on that prestress.
\method demonstrates significant improvements over existing bimanual manipulation techniques across $10$ simulated and $9$ real-world tasks by sizable margins in terms of success rate.
% Besides, the visualization of the task-oriented Gaussian Splatting and the future prediction results provide further interpretation of the proposed bimanual policy.
% a range of challenging tasks in both 
The contributions are as follows:

\begin{itemize}
    \item We propose a general robotic bimanual manipulation agent named \method, which extends prior ManiGaussian by introducing the hierarchical Gaussian world model to learn the multi-body spatiotemporal dynamics for bimanual tasks.

    \item We generate task-oriented Gaussian Splatting to differentiate acting and stabilizing arms for multi-body dynamics modeling, and we propose a hierarchical Gaussian world model with future scene prediction to mine the multi-body dynamics.

    \item We perform comprehensive experiments on $10$ tasks from RLBench2. The results indicate that our method surpasses the state-of-the-art approaches by large relative margins of $131.17$\%. We also evaluate \method in real bimanual settings and obtain $60\%$ success rates across $9$ real-world tasks.
    
\end{itemize}

% We first introduce a task-oriented Gaussian radiance field to label actor and stabilizer arms for multi-body representation. Then, we develop a hierarchical Gaussian world model to predict future scenes, where the leader model incorporates prestress and the follower predicts outcomes based on actor actions to simulate the multi-body interaction.
% Take a deep breath, I want you to act as an experienced academician. I'm writing a scientific paper for CVPR conference. Here is the related work of my draft, please revise it to make it more fluent and compatible with CVPR conference. Only use simple words and expressions. This is very important to my career. """ """

\begin{figure*}[t]
    \centering
    \includegraphics[width=1\textwidth]{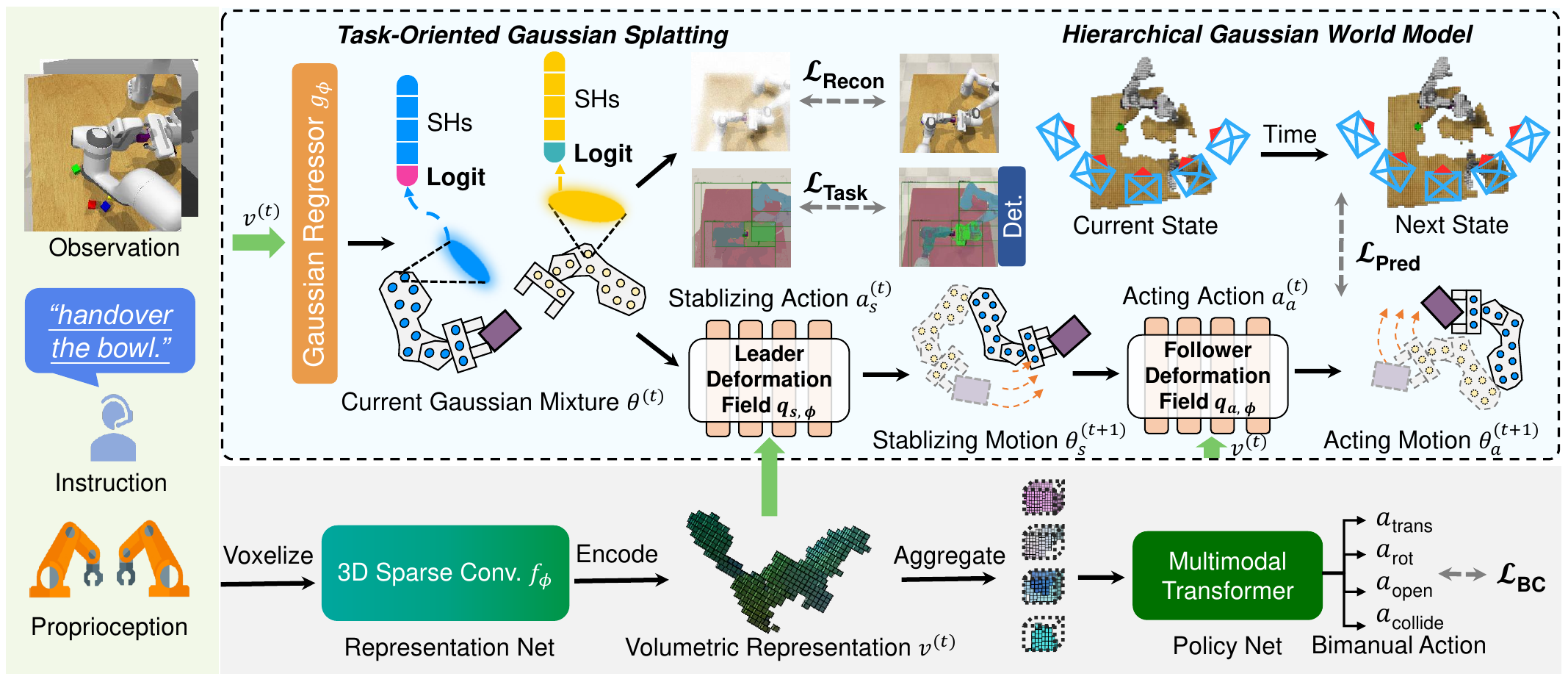}
    \caption{ \textbf{The overall pipeline of \method.} The task-oriented Gaussian radiance assigns unique labels to task-relevant agents and objects, and the hierarchical Gaussian world model upon it predicts future scenes in a leader-follower manner to encode the multi-body dynamics for bimanual manipulation.
    }
    \vspace{-0.6cm}
    \label{fig:pipeline}
\end{figure*}

\section{Related Work}
\label{sec:related_work}

\noindent \textbf{Robotic Bimanual Manipulation.}
Generalizable bimanual manipulation agent enables complex task completion by introducing complex collaboration patterns, which is of great importance in various applications, such as housekeeping \cite{zhang2024empoweringembodiedmanipulationbimanualmobile}, healthcaring \cite{10149474,kim2024surgical}, and manufacturing \cite{buhl2019dual}.
% 双臂预训练模型 （这里把peract2和rdt放在一起说）
Existing methods~\cite{grotz2024peract2benchmarkinglearningrobotic,gbagbe2024bivlavisionlanguageactionmodelbasedbimanual, gao2024bikvilkeypointsbasedvisualimitation,grannen2023stabilizeactlearningcoordinate} attempt to train a foundation model for bimanual manipulation by manual-crafting a large set of demonstrations for imitation learning.
% 但是，双臂采数据直接学很好的策略比较难
However, due to the precise coordination between two high-degree-of-freedom arms required by bimanual tasks, teleoperating demonstrations for training generalizable policies is costly~\cite{tony2023aloha,  ding2024bunnyvisionprorealtimebimanualdexterous, chuang2024activevisionneedexploring, yang2024acecrossplatformvisualexoskeletonslowcost}, which presents challenges for the bimanual manipulation policy model to generalize to unseen tasks. 
% 关键帧比大模型好
% By predicting the keyframe action and leveraging the 3D-aware voxel observation, a recent work PerAct$^2$~\cite{grotz2024peract2benchmarkinglearningrobotic} shows initial potential in single-task bimanual manipulation while notably mitigating the demand of large-scale expert demonstrations.
% 比如PerAct^2系列工作，这是我们的baseline
For instance, by predicting the keyframe action and leveraging the 3D-aware voxel observation, a recent work PerAct$^2$~\cite{grotz2024peract2benchmarkinglearningrobotic} shows initial potential in single-task bimanual manipulation while notably mitigating the demand of large-scale expert demonstrations.
% 但PerAct^2关注policy，忽视representation
% Though PerAct$^2$ improves the sample efficiency in the policy network part, the visual representation that bottlenecks the generalizability of multi-task bimanual manipulation is neglected, which is the main focus of this paper.
Though PerAct$^2$ improves the expressiveness of the policy network part by leveraging scalable multi-modal transformer~\cite{jaegle2021perceiver}, the visual representation that bottlenecks the generalizability of bimanual manipulation is neglected, which is the main focus of this paper.\\
% Therefore, we present to enhance the visual representation with self-
% For example,  PerAct$^2$~\cite{grotz2024peract2benchmarkinglearningrobotic} shows potential in single-task bimanual manipulation by predicting keyframe actions and using 3D-aware voxel observations, reducing the need for large-scale expert demonstrations. While it improves policy expressiveness through a scalable multi-modal transformer~\cite{jaegle2021perceiver}, it overlooks the visual representation bottleneck that limits multi-task bimanual manipulation
\noindent \textbf{Visual Representations for Robot Learning.}
To enhance the generalizability of the robotic manipulation agents, prior arts propose more powerful network architectures~\cite{goyal2024rvt2, gervet2023act3d,shridhar2023peract,grotz2024peract2benchmarkinglearningrobotic} or employ self-supervision~\cite{nair2022r3m,radosavovic2023realmvp,hansen2021soda,lu2025manigaussian} in visual representation learning.
These methods aim to leverage diverse visual observations like mutli-view images~\cite{goyal2023rvt,goyal2024rvt2}, point cloud~\cite{chen2023polarnet, gervet2023act3d}, and voxel~\cite{shridhar2023peract,grotz2024peract2benchmarkinglearningrobotic,liu2024voxactbvoxelbasedactingstabilizing}, but often struggle with limited labeled data.
% However, visual representations usually struggle to emerge enough generalizability only with the limited set of labeled expert demonstrations.
To address this problem, self-supervised methods~\cite{parisi2022pvr,nair2022r3m,radosavovic2023realmvp} enhances generalization ability through auxiliary tasks with prior knowledge.
% 2D自监督训练视觉表征
Earlier studies \cite{parisi2022pvr, nair2022r3m, radosavovic2023realmvp} have focused on enhancing 2D visual representations via self-supervised techniques such as time-contrastive learning~\cite{nair2022r3m} and masked modeling~\cite{radosavovic2023realmvp}, but are limited to simpler tasks.
% These approaches have shown initial success in visual representation learning but are limited to simpler tasks due to a lack of understanding geometric information.
% 大家开始注意到3D，于是开始用NERF
% To tackle more complex manipulation tasks that necessitate 3D scene comprehension, 
% Recent methods \cite{ze2023rl3d, li20223d, zhu2024spa} employ 3D reconstruction techniques like Neural Radiance Fields (NeRFs) \cite{mildenhall2021nerf} and Gaussian Splatting \cite{kerbl20233d}.
% % 随着GS的兴起，开始用GS (ManiGaussian)
% For instance, ManiGaussian \cite{lu2025manigaussian} innovatively encodes scene dynamics through future scene reconstruction in a Gaussian embedding space, demonstrating effectiveness and efficiency in language-conditioned single-arm manipulation tasks \cite{james2020rlbench}.
Recent approaches tackle 3D scene comprehension with techniques like Neural Radiance Fields (NeRFs) \cite{mildenhall2021nerf} and Gaussian Splatting \cite{kerbl20233d}, with method like ManiGaussian \cite{lu2025manigaussian} encoding scene dynamics for single-arm tasks. However, multi-body dynamics in complex tasks require more advanced approaches to handle significant challenges.\\
% 但是双臂多体动力学更难学
% However, the complex multi-body dynamics involved in tasks with multiple agents present significant challenges, resulting in notable performance declines of these methods in bimanual manipulation scenarios. 
% By incorporating hierarchical Gaussian world model to capture the multi-body spatiotemporal dynamics, our method shows improvement 
\noindent \textbf{World Models.}
A world model simulates the future scene according to the current state and agent action, which often serves as an self-supervised objective to encode the underlying scene dynamcis for autonomous agent of various applications like autonomous driving~\cite{wang2023drivedreamer,gao2022enhance}, gaming~\cite{hafner2022deep,hafner2019dream, hafner2023mastering} and robotic manipulation \cite{hansen2023td,wu2023daydreamer}. 
Early research~\cite{ha2018recurrent,hafner2022deep,hafner2019dream,hafner2020mastering} focus on learning a latent space for future predictions with recurrent state-space models, demonstrating significant effectiveness in both simulated and real-world environments. 
% However, as learning latent representations necessitates substantial data, world models that predicts explicit representations , built upon the image ~\cite{du2024learning,seo2022reinforcement,wu2024pre,mendonca2023structured,yang2023learning,bruce2024genie,alonso2024diffusion}, voxel~\cite{zheng2025occworld} and Gaussian domain~\cite{abou2023physically, lu2025manigaussian, zhang2024dynamics} are being attractive.
With the evolution of cutting-edge network architectures, world models that predict high-dimensional representations are often used to predict the future image ~\cite{du2024learning,seo2022reinforcement,mendonca2023structured,seo2023masked}, voxel~\cite{zheng2025occworld} and Gaussian~\cite{abou2023physically, lu2025manigaussian, zhang2024dynamics} domains.
% 
% ManiGaussian~\cite{lu2025manigaussian} first build a Gaussian world model by generalizing the world model to embedding space of dynamic Gaussian Splatting to encode the spatiotemporal dynamics, and achieves initial success in multi-task unimanual manipulation.
However, the multi-agent nature of bimanual manipulation poses challenges to model the mutual interactions between two manipulators and targets, and thus we introduce a hierarchical Gaussian world model to overcome this.\\
% and is limited to simple tasks such as robot control due to the weak representative ability of the implicit features
\noindent \textbf{Gaussian Splatting.}
Gaussian Splatting~\cite{kerbl20233d} is termed by representing scenes using a collection of 3D Gaussian functions that can be `splatted' onto 2D planes with rasterization, compared to implicit models like Neural Radiance Fields (NeRF)~\cite{mildenhall2021nerf,driess2022nerfrl,shim2023snerl}. 
% For a detailed overview, please check the survey~\cite{chen2024survey} on 3D Gaussian Splatting.
% To extend Gaussian Splatting for various complex situations, numerous extensions have been developed. These aim to improve its generalizablity~\cite{szymanowicz2023splatter, zheng2023gps,zou2023triplane,charatan2023pixelsplat,fu2023colmap,xu2024agg,zhang2024gaussiancube}, semantic expressiveness~\cite{qin2023langsplat,zuo2024fmgs,shi2023language,zhou2024feature}, and enable the reconstruction of dynamic scenes~\cite{yang2023real,xie2023physgaussian,yang2023deformable,luiten2023dynamic,wu20234d,liang2023robo360,abou2023physically}.
Recent works~\cite{quach2024gaussian,shorinwa2024splat,lou2024robo, lu2025manigaussian} begin to notice the great potential of the Gaussian radiance field in robotic manipulation, which is able to consistently tracking the manipulator and target with its explicit and editable nature.
Although these methods depicts impressive performance in unimanual settings by incorporating Gaussian-based representation, bimanual manipulation that involves complex multi-body dynamics is still unexplored.
% In this paper, we formulate a dynamic Gaussian Splatting framework to model the scene dynamics of object interactions, which enhances the physical reasoning for agents to complete a wide range of robotic manipulation tasks.

% surgical scene reconstruction~\cite{zhu2024deformable,liu2024endogaussian}

% Take a deep breath, I want you to act as an experienced academician. I'm writing a scientific paper for CVPR conference. Here is the approach section of my draft, please revise it to make it more fluent and compatible with CVPR conference. Only use simple words and expressions. This is very important to my career. """ """

% It consists of a task-oriented Gaussian Splatting and a hierarchical Gaussian world model.

\section{Approach}
\label{sec:approach}
% disentangled generalizable Gaussian Splatting (DG-GS)
% In this section, we first briefly introduce preliminaries on multi-task bimanual manipulation in \Cref{subsec:prelimilaries}, and then we depict an overview of our pipeline in \Cref{subsec:overall}. We present task-oriented Gaussian Splatting for successor multi-body modeling in \Cref{subsec:task_oriented_gs}.
% Subsequently, we introduce the hierarchical Gaussian world model in \Cref{subsec:hierarchical_wm} to encode the multi-body spatiotemporal dynamics.
% The total learning objective is shown in \Cref{subsec:loss}.
In this section, we first briefly introduce preliminaries on bimanual manipulation, and then we depict an overview of our pipeline. Subsequently, we present the task-oriented Gaussian Splatting for multi-body modeling, and introduce the hierarchical Gaussian world model to encode the multi-body spatiotemporal dynamics. Finally, We describe the learning objectives for supervisions.

% which propagates the relevant instance labels provided by the pretrained VLM in the Gaussian radiance field.
% in a generalizable feed-forward manner.

\subsection{Problem Formulation}\label{subsec:prelimilaries}

Language-conditioned bimanual manipulation is essential for next generation's general intelligent robot.
To carry out various bimanual manipulation tasks, the agent must interactively predict the next best poses of both end-effectors to accomplish the human instruction, based on observations including visual input and the robot's proprioception. 
% At each run, the motion to reach the pose is generated by a low-level motion planner, such as RRT-Connect, then the agent receives new observations and begins inference again.
% Each execution involves a low-level motion planner like RRT-Connect to generate the trajectory to the desired pose, then the agent restarts inference with new observations.
The observation at the $t_{th}$ time step is  $\mathbf{o}^{\left ( t \right ) } = ( C^{\left ( t \right ) } ,D^{\left ( t \right ) },P^{\left ( t \right ) })$, where  $C^{\left ( t \right ) }$ and $D^{\left ( t \right ) }$ respectively represent the RGB image and the depth image. 
$P^{\left ( t \right ) }$ is the proprioception that contains current time and gripper states.
The action $\mathbf{a}^{(t)}$ for each end-effector at the $t_{th}$ step contains the position $a^{(t)}_{\text{trans}}\!\in\! \mathbb{R}^{\!100^3}$, orientation $a_{\text {rot}}^{(t)}\!\in\! \mathbb{R}^{\!(360\! /\! 5) \!\times\! 3}$, openness $a^{(t)}_{\text{{open\phantom{t}}}}\!\in\! [0,\!1\!]$ and an indicator of whether to invoke collision avoidance of the motion planner $a^{(t)}_{\text{collide}}\!\in\! [0,\!1\!]$.
% a limited set of
% $\mathcal{D} \!=\! \left\{(\mathbf{o}^{(1)}, \mathbf{a}_{\text{s}}^{(1)}, \mathbf{a}_{\text{a}}^{(1)}), ..., (\mathbf{o}^{(K)}, \mathbf{a}_{\text{s}}^{(K)}, \mathbf{a}_{\text{a}}^{(K)})\right\}$
For robot learning, we assume access to $K$ offline trajectories composed of observation, left-arm action and right-arm action triplets paired with human instructions.
% , where $\mathbf{a}_{\text{s}}^{(t)}$ and $\mathbf{a}_{\text{a}}^{(t)}$ respectively refer to the actions for left and right end-effectors. 
% To mitigate the constraint of the limited demonstrations, traditional arts attempt to enhance the visual representation by reconstructing consistent 3D geometry and predicting the future observation via self-supervised learning. 
To address the issue of limited demonstrations, traditional arts~\cite{lu2025manigaussian} enhance the visual representation by mining the spatiotemporal dynamics via self-supervised future scene reconstruction.
However, the multi-body nature of the bimanual manipulation poses challenges to modeling the spatiotemporal dynamics precisely, and the decoded actions based on the ineffective visual representation fail to complete human instructions with incorrect collaboration patterns.
% single-view

\subsection{Overall Pipeline}\label{subsec:overall}

The overall pipeline of our \method method is shown in \Cref{fig:pipeline}. 
% 解释为啥要用stablizing and acting
We first develop a task-oriented Gaussian radiance field to distinguish stabilizing and acting arms, where a stabilizing arm secures the object in hand while an acting arm performs the task to mitigate the multi-modality of bimanual manipulation.
% for multi-body spatiotemporal dynamics modeling, where 
Subsequently, we build a hierarchical Gaussian world model in leader-follow architecture upon it to forecast the future scene, through which the multi-body spatiotemporal dynamics can be encoded to discover complex collaboration patterns.
%%%%%%%%%%%%%%%%%%%%%%%%%%%%%%%%%%%%%%%%%%%%%%
More specifically, we transform the visual input from RGB-D cameras to voxel space based on the calibrated camera parameters, which is then encoded by a sparse convolutional network as the visual representation in a volumetric format.
For task-oriented Gaussian Splatting, we design a feed-forward Gaussian regressor to infer the task-oriented Gaussian radiance field from the visual representation, where each Gaussian particle is assigned with a task-oriented instance logit distilled from pretrained VLMs to differentiate acting and stabilizing arms for multi-body interactions. 
% dealing with the various environments in multi-task bimanual manipulation.
% For the hierarchical Gaussian world model, we introduce a leader and a follower deformation field to take the multi-body spatiotemporal dynamics into account, which first outputs the prestress provided by the stabilizer arm and concludes the final movement determined by the actor arm in order.
We create a hierarchical Gaussian world model with a leader-follower architecture to learn visual representation through future scene prediction. The leader anticipates deformation from the motion of the stabilizing arm, allowing the follower to generate the effects of the movement of the acting arm.
Finally, we employ multi-modal transformer PerceiverIO~\cite{jaegle2021perceiver} to predict the optimal robot actions based on the enhanced volumetric representation, which comprehends the multi-body spatiotemporal dynamics and thus can complete human instructions with precise collaboration patterns.

% for Bimanual Manipulation
\subsection{Task-Oriented Gaussian Splatting }\label{subsec:task_oriented_gs}
% 为啥要提出Task-Oriented Gaussian Splatting
In order to capture the multi-body spatiotemporal dynamics for general bimanual manipulation tasks, we start with a task-oriented Gaussian Splatting that disentangles the acting and stabilizing arms from cluttered scenes for representing the visual scene.
% 简要介绍原始GS
Gaussian Splatting \cite{kerbl20233d} is trending with its explicit nature that enables rapid rendering via rasterization. 
A Gaussian radiance field represents a scene with multiple Gaussian primitives, which can be parameterized by $\theta_i = (\mu_i, c_i, r_i, s_i, \sigma_i)$, which respectively represent the position, color, orientation, scaling, and opacity for the $i$-th Gaussian primitive.
To render a novel image $C$, those 3D Gaussian primitives can be projected onto the 2D camera plane via differential tile-based rasterization. In this process, a typical pixel $\mathbf{p}$ can be colored by the alpha-blend rendering:
\begin{equation}\label{eq:rgb_rendering}
    C(\mathbf{p}) =\! \sum_{i=1}^{N} \alpha_i c_i \prod_{j=1}^{i-1} (1-\alpha_j)
\end{equation}
where $N$ is the number of Gaussians in a tile, $\alpha_i$ represents the 2D density of the Gaussian points that can be computed by $\mu_i$, $r_i$ and $s_i$ of the parameters.
% baseline (原始Gaussian) 不能建模实例级别
Though the vanilla Gaussian Splatting shows effectiveness in reconstructing 3D appearance and geometry, it struggles to generate high-quality task-oriented labels for relevant instances, which is of significance to focus on learning precise multi-body dynamics for bimanual manipulation tasks.
% 我们修改GS的参数，让instance和RGB等价，从而可以实现原始GS实现不了的区分stablizing和acting arm
To this end, we modify the Gaussian parameters and parameterize a Gaussian regressor to construct a task-oriented Gaussian radiance field, where the instance labels that distinguish the stabilizing and acting manipulators are learned simultaneously with the appearance and geometry. 
% 在这里就加上时间，写出带时间和实例的完整公式，因为不想再写一遍这个公式
% Besides, to effectively model the spatiotemporal dynamics during task completion via the Gaussian points, we formulate a time-variant Gaussian Splatting field, where the parameter of the $i_{th}$ Gaussian at the $t_{th}$ is:
Besides, we enable the Gaussian particles to move with discrete time to account for the spatiotemporal dynamics of the scene, where the parameter of the $i_{th}$ Gaussian at the $t_{th}$ is:
% \begin{equation}\label{eq:dynamic_gaussian_Splatting}
%     \theta_i^{(t)}=\{(\mu_{i}^{(t)}, c_i^{(t)}, r_{i}^{(t)}, s_i^{(t)}, \sigma_i^{(t)}, l_i^{(t)})\}_{i=0}^N.
% \end{equation}
\begin{equation}\label{eq:dynamic_gaussian_Splatting}
    \theta_i^{(t)}=(\mu_{i}^{(t)}, c_i^{(t)}, r_{i}^{(t)}, s_i^{(t)}, \sigma_i^{(t)}, l_i^{(t)}).
\end{equation}
The positions, colors, orientations, scales, and opacities with the superscript $t$ represent their counterparts at the $t_{th}$ step in the movement.
We append a $l_i^{(t)}\in \mathbb{R}^3$ variable as the instance-level logit to the Gaussian parameter set, which represents the probability of the Gaussian point belonging to a specific task-relevant instance including different manipulators or target objects.
% 怎么实现
The instance map can also be rendered by projecting the instance-level logits of Gaussian primitives to the 2D camera plane.
%so that the semantic feature can be propagated according to the reconstructed appearance and geometry to compensate for the lack of ground-truth instance map.
To be specific, the expected instance logits $L$ of a typical pixel $\mathbf{p}$ can be written as:
% mask光栅化
\begin{flalign}
L(\mathbf{p})=\!\sum_{i=1}^{N}\alpha_{i}l_{i}\prod_{j=1}^{i-1}(1-\alpha_{j}),
\end{flalign}
where we omit the $t$ superscript for brevity. To obtain the ground-truth instance map for both manipulators and target objects, we prompt the pretrained VLMs such as the open-vocabulary detector GroundedSAM~\cite{ren2024grounded} based on the keywords from human instructions. 
% To differentiate the stabilizing and acting arms, we also incorporate two prompts `left arm' and `right arm', and then filter out any generated masks whose confidence level is below a threshold.

% \begin{flalign}
% % 用s表示语义类别还是label（维度3）
% \theta_{i}^{(t)}=(\mu_{i}^{(t)},c_{i}^{(t)},r_{i}^{(t)},l_{i}^{(t)},\sigma_{i}%
% ^{(t)},f_{i}^{(t)},s_{i}^t).
% \end{flalign}

\subsection{Hierarchical Gaussian World Model for Bimanual Manipulation}\label{subsec:hierarchical_wm}

% 动机
Though the multi-body nature of bimanual manipulation enables complicated task completion, it also introduces novel challenges for the world model to learn multi-body spatiotemporal dynamics beyond unimanual manipulation.
% 强调hierarchical
In order to capture the multi-body spatiotemporal dynamics for the visual representation in general bimanual manipulation tasks, we propose a hierarchical Gaussian world model with a leader-follower architecture for precise future scene prediction. The leader predicts Gaussian Splatting deformation caused by motions of the stabilizing arm, through which the follower generates the physical consequences resulted from the movement of the acting arm.
% 细节
The hierarchical Gaussian world model models the movement of explicit Gaussian points conditioned on the robot's action.
For future prediction of our hierarchical Gaussian world model, the dominant movement can be regarded as rigid-body transform. Therefore, we only predict the SE(3) movement of Gaussian particle following the Newton-Euler equation, while keeping the inherent properties including color, scaling, opacity, and instance logits the same along the Markovian transition.
The Gaussian particle changes caused by arm motions can be formulated as:
\begin{flalign}\label{eq:propagation}
\small
(\boldsymbol{\mu}^{\!(t\!+\!1)}\!,\!\mathbf{r}^{\!(t\!+\!1)})\!=\!(\boldsymbol{\mu}^{\!(t)}\!+\!\Updelta\boldsymbol{\mu}_{\text{s}}^{\!(t)}\!+\!\Updelta\boldsymbol{\mu}_{\text{a}}^{\!(t)}\!,\!\mathbf{r}^{\!(t)}\!+\!\Updelta \mathbf{r}_{\text{s}}^{\!(t)}\!+\!\Updelta \mathbf{r}_{\text{a}}^{\!(t)}),
\end{flalign}
% \begin{equation}\label{eq:propagation}
%     \theta_i^{(t+1)}\!=\!(\mu_{i}^{(t)}\!+\!\Updelta\mu_{i}^{(t)}\!+\!\Updelta\mu_{i}^{(t)}, c_i^{(t)}, r_{i}^{(t)}\!+\!\Updelta r_{i}^{(t)}\!+\!\Updelta r_{i}^{(t)}, s_i^{(t)}, \sigma_i^{(t)}, l_i^{(t)}).
% \end{equation}
% \begin{equation}\label{eq:propagation}
%     \theta_i^{(t+1)}\!=\!(\mu_{i}^{(t)}\!+\!\Updelta\mu_{i}^{(t)}, r_{i}^{(t)}\!+\!\Updelta r_{i}^{(t)}) \cup (c_i^{(t)}, s_i^{(t)}, \sigma_i^{(t)}, l_i^{(t)}).
% \end{equation}
where we denote the changes of positions and orientation by $\Updelta \boldsymbol{\mu}_{\text{s}}^{(t)}$, $\Updelta \boldsymbol{r}_{\text{s}}^{(t)}$ and $\Updelta \boldsymbol{\mu}_{\text{a}}^{(t)}$, $\Updelta \boldsymbol{r}_{\text{a}}^{(t)}$ for stabilizing and acting manipulator, respectively. The subscript $i$ for each Gaussian particle is omitted here for brevity.
% 简要介绍overall flow，这里只介绍输入输出？ (parameterize)
To predict the movements described in \Cref{eq:propagation} for multi-body spatiotemporal dynamics learning, we parameterize a hierarchical Gaussian world model that takes the visual representation as input, and outputs the future multi-view images for photometric supervision.
% 细节
More specifically, the hierarchical Gaussian world model contains a representation network $f_{\phi}$ that infers intermediate visual representation from the voxel observation, where $\phi$ refers to the learnable parameters. Then, a Gaussian regressor $g_{\phi}$ is utilized to reconstruct the current task-oriented Gaussian radiance field in a feed-foward manner.
For future scene reconstruction, a leader deformation model $q_{\text{s}, \phi}$ interprets the preliminary movements imposed by the stabilizing arm as a Gaussian deviation $\theta_{\text{s}}^{\left ( t\!+\!1 \right ) }$, a follower deformation model $q_{\text{a}, \phi}$ predicts the physical consequences $\theta_{\text{a}}^{\left ( t\!+\!1 \right ) }$ by concluding both the acting and stabilizing arm. At last, a Gaussian renderer $\mathcal{R}$ in \Cref{eq:rgb_rendering} projects the predicted Gaussian radiance field onto the 2D camera plane:
\begin{equation}
\small
\left\{
\begin{array}{l}
\text{Representation: } v^{\left ( t \right ) } \!=\! f_{\phi}( o^{\left ( t \right ) } ) \\ 
\text{Gaussian regressor: } \theta^{\left ( t \right ) } \!=\! g_{\phi}( v^{\left ( t \right ) } ) \\
\text{Leader model: }  \theta_{r}^{\left ( t\!+\!1 \right ) } \!=\! q_{\text{s}, \phi}(\theta^{\left ( t \right ) },\mathbf{a}_{\text{s}}^{(t)},v^{(t)} ) \\  
\text{Follower model: } \theta_{l}^{\left ( t\!+\!1 \right ) } \!=\! q_{\text{a}, \phi}(\theta_{r}^{\left ( t\!+\!1 \right )}\!,\mathbf{a}_{\text{s}}^{(t)}\!,\mathbf{a}_{\text{a}}^{(t)}\!,v^{(t)} ) \\
\text{Gaussian renderer: } C^{\left ( t\!+\!1 \right ) }, L^{\left ( t\!+\!1 \right ) } \!=\! \mathcal{R}(\theta^{\left ( t\!+\!1 \right ) }),
\end{array}
\right.    
\end{equation}
% \Updelta
where $v^{\left ( t \right ) }$ denotes the enhanced visual representation, $\mathbf{a}_{\text{s}}^{(t)}$ and $\mathbf{a}_{\text{a}}^{(t)}$ are the stabilizing and acting actions at the $t_{th}$ step.
In order to effectively forecast future scenes with the hierarchical Gaussian world model, the visual representation is enhanced to capture multi-body spatiotemporal dynamics of the environment, which is crucial for generating appropriate bimanual actions with complex collaboration patterns.

% 解释符号

% $w$ is the camera pose for the view where we project the Gaussian primitives. 

\begin{table*}[t]
\centering
\caption{ \textbf{Multi-Task Test Results.} Mean success rates (\%) of multi-task agents trained with 100 demonstrations per task and evaluated over 25 episodes.}
\label{table:comparison_with_sota}
\setlength{\tabcolsep}{10pt}
\small
\renewcommand{\arraystretch}{1}%
\begin{tabular}{l|cccccc}
\toprule
\textbf{Method / Task} & \makecell{\texttt{pick} \\ \texttt{laptop}} & \makecell{\texttt{straighten} \\ \texttt{rope}} & \makecell{\texttt{lift} \\ \texttt{tray}} & \makecell{\texttt{push} \\ \texttt{box}} & \makecell{\texttt{handover} \\ \texttt{easy}} & \makecell{\texttt{put in} \\ \texttt{fridge}} \\
\midrule
% PerAct~\cite{shridhar2023peract}-LF & 0 & 4 & 8 & 12 & 4 & 0 \\
% PerAct$^2$~\cite{grotz2024peract2benchmarkinglearningrobotic} & \textbf{8} & 16 & 4 & \textbf{40} & 4 & 0 \\
PerAct$^2$~\cite{grotz2024peract2benchmarkinglearningrobotic} & \textbf{12} & 24 & 1 & 6 & \textbf{41} & 3 \\
ManiGaussian~\cite{lu2025manigaussian} & \underline{8} & \underline{28} & \underline{4} & \underline{24} & 36 & \underline{4} \\
\rowcolor{\ourcolor}\textbf{\method (Ours)} & \textbf{12} & \textbf{40} & \textbf{8} & \textbf{48} & \underline{40} & \textbf{28}\\
\midrule
\textbf{Method / Task}  & \makecell{\texttt{press} \\ \texttt{buttons}} & \makecell{\texttt{handover} \\ \texttt{item}} & \makecell{\texttt{sweep to} \\ \texttt{dustpan}} & \makecell{\texttt{take out} \\ \texttt{tray}} & \makecell{\textbf{Average} \\ \textbf{Success} $\uparrow$} & \makecell{\textbf{Average} \\ \textbf{Rank} $\downarrow$} \\
\midrule
% PerAct~\cite{shridhar2023peract}-LF & 8 & 0 & 0 & 0 & 1.6 & 2.9 \\
% PerAct$^2$~\cite{grotz2024peract2benchmarkinglearningrobotic} & 12 & 0 & 0 & 0 & 2.4 & 1.8 \\
PerAct$^2$~\cite{grotz2024peract2benchmarkinglearningrobotic} & \underline{47} & 11 & 0 & 9 & 15.4 & 2.5 \\
ManiGaussian~\cite{lu2025manigaussian} & 36 & \underline{12} & \underline{24} & \underline{12} & \underline{18.8} & \underline{2.2} \\
\rowcolor{\ourcolor}\textbf{\method (Ours)} & \textbf{48} & \textbf{20} & \textbf{92} & \textbf{16} & \textbf{35.6} & \textbf{1.1} \\
\bottomrule
\end{tabular}
\end{table*}

\subsection{Learning Objectives}\label{subsec:loss}

% \noindent\textbf{Current Scene Consistency Loss.}
\noindent\textbf{Current Scene Reconstruction.}
% Reconstructing the current scene based on the current Gaussian parameters accurately can enhance the performance of the Gaussian regressor. To achieve this goal, we introduce the consistency objective between the realistic current observation and the rendered according to the current Gaussian parameters:
% 动机 (多体时空动力学里面的'空')
To encode the spatial consistency into the visual representation for further multi-body spatiotemporal dynamics digesting, we impose a multi-view photometric loss to regularize the current Gaussian Splatting generated from the Gaussian regressor:
% 跟原始GS重建loss的区别
% The current Gaussian radiance field is constructed by the Gaussian regressor in a feed-forward manner.
% To supervise the reconstructed 3D appearance, rather than using a combination of $\ell_1$ loss and D-SSIM loss in the original Gaussian Splatting paper~\cite{kerbl20233d}, we alternatively utilize a simpler MSE loss function for the reconstruction objective:    % 不强调了
% \begin{flalign}
% \mathcal{L}_{\text{Recon}}=\|\mathbf{C}^{(t)}-\hat{\mathbf{C}}^{(t)}\|_{2}^{2},
% \end{flalign}
% \begin{flalign}\label{eq:objective_rgb}
% \mathcal{L}_{\text{Recon}}=\|C^{(t)}-\hat{C}^{(t)}\|_{2}^{2},
% \end{flalign}
\begin{flalign}\label{eq:objective_rgb}
\mathcal{L}_{\text{Recon}}=\sum_{\mathbf{p}} \|C^{(t)}(\mathbf{p})-\hat{C}^{(t)}(\mathbf{p})\|_{2}^{2},
\end{flalign}
where $C^{(t)}$ and $\hat{C}^{(t)}$ stand for the predicted and ground-truth 2D images from a randomly selected view at $t_{th}$ time step, respectively.
% $\mathbf{C}^{(t)}$ and $\hat{\mathbf{C}}^{(t)}$

% \noindent\textbf{Task-relevant Instance Consistency Loss.}
% \noindent\textbf{Task-oriented Gaussian Splatting.}
% \noindent\textbf{Task-oriented Reconstruction.}
\noindent\textbf{Task-oriented Embodiment Mask Prediction.}
% 回顾动机和实现方式
The task-oriented Gaussian radiance field differentiates acting and stabilizing arms for the follow-up multi-body spatiotemporal dynamics modeling.
% which predicts high-level instance labels including different arms and target objects for each Gaussian particle.
To optimize this, we first aggregate the logits embedded in the Gaussian particles by rendering them to the camera plane via rasterization, and then implement a cross-entropy objective:
% \begin{equation}\label{eq:objective_task}
% \mathcal{L}_{\text{Task}}=\|\mathbf{L}^{(t)}-\hat{\mathbf{L}}^{(t)}\|_{2}^{2},
% \end{equation}
% \begin{equation}\label{eq:objective_task}
% \mathcal{L}_{\text{Task}}=-\sum_{\mathbf{r} \in \mathcal{R}}\left[\sum_{l=1}^L p^l(\mathbf{r}) \log \hat{p}_c^l(\mathbf{r})\right]
% \end{equation}
% \sum_{l=1}^L
\begin{equation}\label{eq:objective_task}
\mathcal{L}_{\text{Task}}=-\sum_{\mathbf{p}}\sum_{l} \hat{B}^l(\mathbf{p}) \log B^l(\mathbf{p})
\end{equation}
where $B^l$ and $\hat{B}^l$ are the predicted and ground-truth probability. The predicted probability is computed by normalizing the rendered instance logit map $L$ via softmax, while the ground-truth probability is a discrete label obtained by prompting the VLM of a specific label $l$ in the 2D camera plane. 
% To avoid optimization collapse, we also add the task-irrelevant background as an additional label, which is not the manipulators nor the target.

% \noindent\textbf{Future Scene Consistency Loss.}
\noindent\textbf{Future Scene Prediction.}
To embed the multi-body spatiotemporal dynamics in the visual representations, we encourage the predicted scene based on the learned Gaussian parameters to get close to the ground-truth one.
As we can not directly access the ground-truth future Gaussian radiance field, the training goal is to align predicted future images from multiple views with the ground-truth images obtained by actually taking the bimanual action, as follows:
\begin{flalign}
 % 未来场景预测
\mathcal{L}_{\text{Pred}} = \|\hat{\mathbf{C}}^{(t+1)}-\mathbf{C}^{(t+1)}\|_{2}^{2}
\end{flalign}
where $\mathbf{C}^{(t+1)}$ and $\hat{\mathbf{C}}^{(t+1)}$ represent the predicted and ground-truth future image, respectively. 
% In order to effectively forecast future scenes, the visual representation must capture the physical characteristics of the environment, which is crucial for generating appropriate bimanual actions with complex collaboration patterns.

% (a^{(t)}, o^{(t)})

\noindent\textbf{Behavior Cloning.}
% The distribution parameters in our dynamic Gaussian framework are leveraged to predict the optimal action of the robot arm and grippers for general manipulation tasks. 
Following~\cite{shridhar2023peract, grotz2024peract2benchmarkinglearningrobotic} for fair comparisons, we leverage a multi-modal transformer PerceiverIO~\cite{jaegle2021perceiver} to select the best candidates from discretized action bins based on enhanced volumetric representation and language instruction, where we leverage the cross-entropy loss $CE$ to optimize action prediction as a classification problem:
% Action prediction Loss
% \begin{flalign}
% \mathcal{L}_{\text{  BC}}=CE(p_{\text{  trans}},p_{\text{  rot}},p_{\text{  open}},p_{\text{  col}}),
% \end{flalign}
% \begin{equation}
% \mathcal{L}_{\text {BC}}\!=\!CE(p^{\text{left}}_{\text{trans}}, p^{\text{left}}_{\text{rot}}, p^{\text{left}}_{\text{open}}, p^{\text{left}}_{\text{col}}) + CE(p^{\text{right}}_{\text{trans}}, p^{\text{right}}_{\text{rot}}, p^{\text{right}}_{\text{open}}, p^{\text{right}}_{\text{col}})
% \end{equation}
\begin{equation}
\mathcal{L}_{\text {BC}}\!=\!CE(\mathbf{a}^{(t)}_{\text{left}}, \hat{\mathbf{a}}^{(t)}_{\text{left}}) + CE(\mathbf{a}^{(t)}_{\text{right}}, \hat{\mathbf{a}}^{(t)}_{\text{right}}),
\end{equation}
where $\hat{\mathbf{a}}^{(t)}_{\text{left}}$ and $\hat{\mathbf{a}}^{(t)}_{\text{right}}$ are ground-truth actions of the left and right manipulators from provided expert demonstrations, respectively.
The overall objective for our \method agent at each time step is written as:
\begin{equation}
\mathcal{L}=\mathcal{L}_{\text{BC}} + \lambda_{\text{  Recon}}\mathcal{L}_{\text{Recon}}+\lambda_{\text{  Task}} \mathcal{L}_{\text{Task}}+\lambda_{\text{Pred}} \mathcal{L}_{\text{Pred}},
\end{equation}
where the hyperparameters $\lambda_{\text{Recon}}, \lambda_{\text{Task}}, \lambda_{\text{Pred}}$ can be tuned to balance various objectives. 
\newcommand{\tabref}[1]{Table~\ref{#1}}

\begin{table*}[t]
  \centering
  % \scriptsize
  \small
    \caption{ \textbf{Comparison of Our Methods with Different Techniques.} We manually categorize the $12$ RLBench2 task to $3$ groups for further interpretability, then we select one task from each category. For more details, please refer to the supplementary file.}
  \setlength{\tabcolsep}{6pt} 
    % \begin{tabular}{c|ccc|cccccc|c}
    \begin{tabular}{c|ccc|ccc|c}
    \toprule
    % Row ID & Gaussian Splatting & Task-oriented GS & Hierarchical GWM & \texttt{Long} & \texttt{Planning} & \texttt{Tools} & \texttt{Motion} & \texttt{Lift} & \texttt{Occlusion} & \textbf{Average} \\
    \textbf{Row ID} & \textbf{Gaussian Splatting} & \textbf{Task-oriented GS} & \textbf{Hierarchical GWM} & \makecell{\texttt{sweep to} \\ \texttt{dustpan}} & \makecell{\texttt{handover} \\ \texttt{item}} & \makecell{\texttt{push} \\ \texttt{box}}  & \textbf{Average} $\uparrow$ \\
    \midrule
    1 & \no & \no & \no & 0 & 11 & 6 & 5.67\\
    2 & \yes & \no & \no & 24 & 12 & 24   & 20.00\\
    3 & \yes & \yes & \no & \underline{32} & \underline{16} & \underline{32} & \underline{26.67}\\
    \rowcolor{\ourcolor}
    4 & \yes & \yes & \yes & \textbf{92} & \textbf{20} & \textbf{48} & \textbf{60.00}\\
    \bottomrule
    \end{tabular}
  \label{table:ablation_study}
\end{table*}

\section{Experiments}
\label{sec:experiments}

% We first introduce the simulated and real-world experiment setup in \Cref{subsec:setup}. Then, we report the simulated performance of our method compared with the state-of-the-art approaches in \Cref{subsec:comp_with_sota}. We conduct a comprehensive ablation study to validate the proposed task-oriented Gaussian radiance field and the hierarchical Gaussian world model in \Cref{subsec: ablation_study}. We interpret the proposed techniques by visualization in \Cref{subsec:qualitative}. 
% Finally, we present qualitative results to depict the effectiveness of our \method in real-world settings in \Cref{subsec:real_world_exp}.
% More results and case studies can be found in the supplementary file.

In this section, we first introduce the simulated and real-world experiment setups. Then, we report the simulated performance of our method compared with the state-of-the-art approaches. We conduct a comprehensive ablation study to validate the proposed task-oriented Gaussian radiance field and the hierarchical Gaussian world model. We interpret the proposed techniques by visualization. 
Finally, we present qualitative results to depict the effectiveness of our \method in real-world settings.
% More results and case studies can be found in the supplementary file.

\begin{figure*}[t]
    \centering
    \includegraphics[width=1\textwidth]{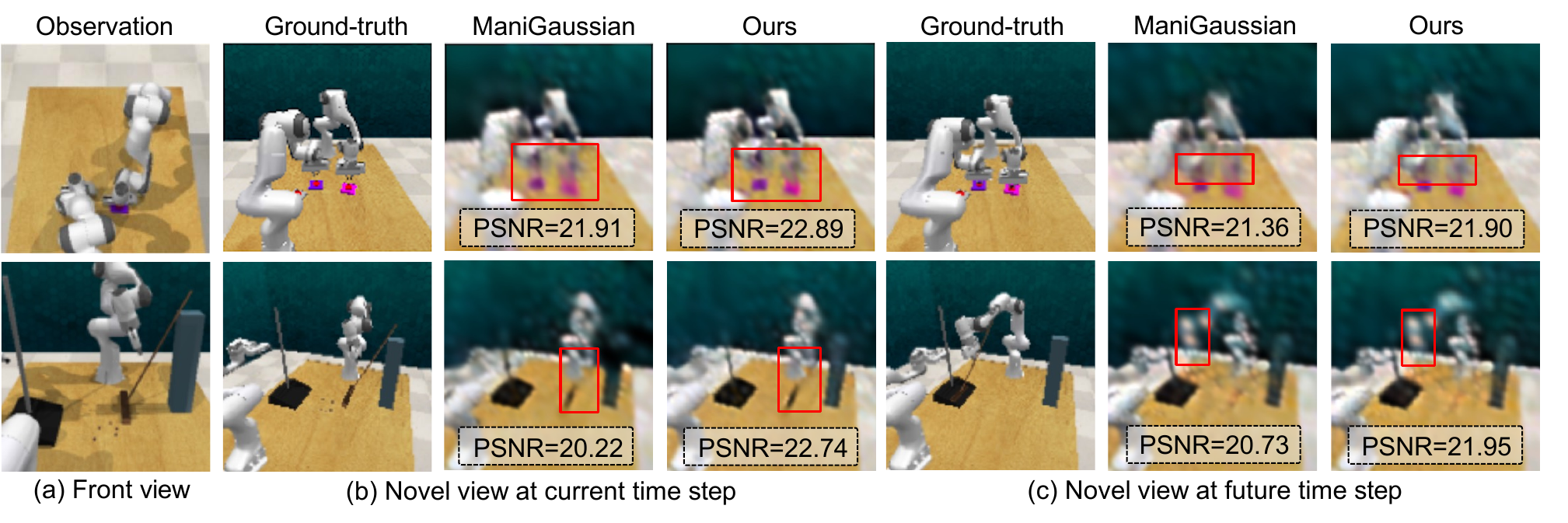}
    \caption{ \textbf{Novel View Synthesis Results.} Our \method captures the multi-body spatiotemporal dynamics precisely, while ManiGaussian fails to model it. Note that we turn off the behavior cloning loss for better illustration.}
    \vspace{-0.6cm}
    \label{fig:qualitative_novel_view_synthesis}
\end{figure*}

% \begin{table*}[t]
% \centering
% \caption{\small \textbf{Real-world Results.} We train and evaluate \method on $9$ challenging real-world tasks.}
% \label{table:real_world_results}
% \setlength{\tabcolsep}{3.5pt}
% \small
% \renewcommand{\arraystretch}{1}%
% \begin{tabular}{l|ccccc}
% \toprule
% \textbf{Task} & \texttt{Lift Box} & \texttt{Fold Clothes} & \texttt{Pick \& Place Sync} & \texttt{Handover Bowl} & \texttt{Press Handsan} \\
% \midrule
% ManiGaussian~\cite{lu2025manigaussian} & 80 & 60 & 70 & 20 & 20 \\
% \midrule
% \rowcolor{\ourcolor}
% \textbf{\method} & 100 & 80 & 80 & 20 & 20 \\
% \midrule
% \textbf{Task} & \texttt{Ping Pong} & \texttt{Relocate Brush} & \texttt{Push Buttons} & \texttt{Pick \& Place Async} & \textbf{Average} \\
% \midrule
% ManiGaussian~\cite{lu2025manigaussian} & 60 & 0 & 40 & 60 & 46 \\
% \midrule
% \rowcolor{\ourcolor}
% \textbf{\method} & 100 & 0 & 60 & 80 & 60 \\
% \bottomrule
% \end{tabular}
% \end{table*}

% \begin{figure*}[t]
% \begin{figure}[t]
%     \centering
%     \includegraphics[width=0.49\textwidth, height=5cm]{pic/real-world-rollout.pdf}
%     \caption{Rollouts of \method and the previous state-of-the-art robotic bimanual manipulation method PerAct$^2$.
%     }
%     \vspace{-0.2cm}
%     \label{fig:real_world_gallery}
% \end{figure}
% \end{figure*}

\subsection{Experiment Setup}\label{subsec:setup}

\noindent\textbf{Simulation.} 
For benchmarking, we conduct our simulation experiments on  RLBench2~\cite{grotz2024peract2benchmarkinglearningrobotic}, a bimanual extension from the popular RLBench \cite{james2020rlbench}. It contains $10$ challenging languaged-conditioned manipulation tasks varying from different challenge levels. 
% To achieve high success rates across multiple tasks, the robot is required to comprehend the multi-body spatiotemporal dynamics for precise collaboration, rather than just imitating expert demonstrations. 
For agent observation, we employ RGB-D images from six cameras 
% (i.e. front, left, right, wrist left, wrist right, overhead) 
with a resolution of $256 \times 256$, in line with~\cite{grotz2024peract2benchmarkinglearningrobotic}. We use the same number of cameras as ManiGaussian~\cite{lu2025manigaussian} to provide multi-view supervision for fair comparisons. In the training phase, we provide $100$ demonstrations for each task, which are generated by an Oracle scripted expert. 

\noindent\textbf{Real Robot.} 
The experimental setup consists of two Universal Robots UR5e arms equipped with Robotiq 2F-85 grippers, controlled via two Xbox controllers to collect demonstration data. Two RGB-D Realsense cameras capture $640\times480$ resolution images at $30$ Hz. While multi-view cameras are utilized during the training phase, only a single camera is used during inference. 
% The easy\_handeye package is used for camera-to-robot calibration. 
We collect $30$ real-world human demonstrations for training, while evaluating the trained policy for $10$ episodes with a Nvidia RTX 4080 GPU.
% For more setup details, please refer to the supplementary file.

\noindent\textbf{Baselines.} 
We compare our \method with the state-for-the-art robotic bimanual manipulation methods, including PerAct$^2$~\cite{grotz2024peract2benchmarkinglearningrobotic}, which is a strengthened version of the widely-used single-arm agent PerAct~\cite{shridhar2023peract}.
% Additionally, we include PerAct-LF in our comparisons, which employs a leader-follower \cite{grotz2024peract2benchmarkinglearningrobotic} architecture in the policy network to consider the stabilizing and acting roles in bimanual manipulation.
Additionaly, we include the former version Manigaussian \cite{lu2025manigaussian} by modifying the action dimension to be compatible with bimanual settings. 
The primary evaluation metric is the task success rate, which is calculated as the percentage of episodes in which the agent completes the task within a budget of $25$ steps.

% \noindent\textbf{Implementation Details.} 
% Following~\cite{shridhar2023peract,grotz2024peract2benchmarkinglearningrobotic}, we employ  SE(3) augmentation to the training set to enhance the robustness of the compared methods. 
% by perturbing the voxel observation with $[\pm0.125m,\pm0.125m,\pm0.125m]$ translation, and rotate it around the $z$-axis by $[0^{\circ},0^{\circ},45^{\circ}]$. 
% All compared methods are trained on two NVIDIA RTX 3090 GPUs for $100$k iterations with a batch size of 2 for fair comparisons. We utilize the LAMB optimizer \cite{you2020largebatchoptimizationdeep} within a constant learning rate of $5\times10^{-4}$ to update the model parameters. Additionally, we insert two warm-up phases in the first 2k steps before optimizing \Cref{eq:objective_rgb} and 3k steps before optimizing \Cref{eq:objective_task} for better convergence.
% To mitigate the impact of parameter size, we utilize the same version of PerceiverIO as the action decoder across all baselines. 
% cosine scheduler with

\subsection{Comparisons with the State-of-the-Arts}\label{subsec:comp_with_sota}

We compare the proposed \method with the state-of-the-art methods in the commonly-used bimanual manipulation benchmark RLBench$^2$ and report the performances.
Our \method achieves the best performance across $10$ tasks ranging in different challenging levels, which demonstrates the superiority of the proposed techniques.
Notably, by digesting the multi-body spatiotemporal dynamics in bimanual manipulation tasks, \method outperforms its former version ManiGaussian by a sizable relative improvement of $89.36$\% ($18.8$\% \text{vs} $35.6$\%). 
Even the enhanced version of PerAct$^2$ that leverages six cameras to ensure seamless observation is also defeated by the proposed \method, underling the importance of mining the consistency from the provided multi-view images and the multi-body spatiotemporal dynamics.
% underling the importance of mining the consistency from the provided multi-view images.
The results prove the capacity of the proposed \method to handle general robotic manipulation tasks.

\begin{figure*}[t]
    \centering
    \includegraphics[width=0.9\textwidth]{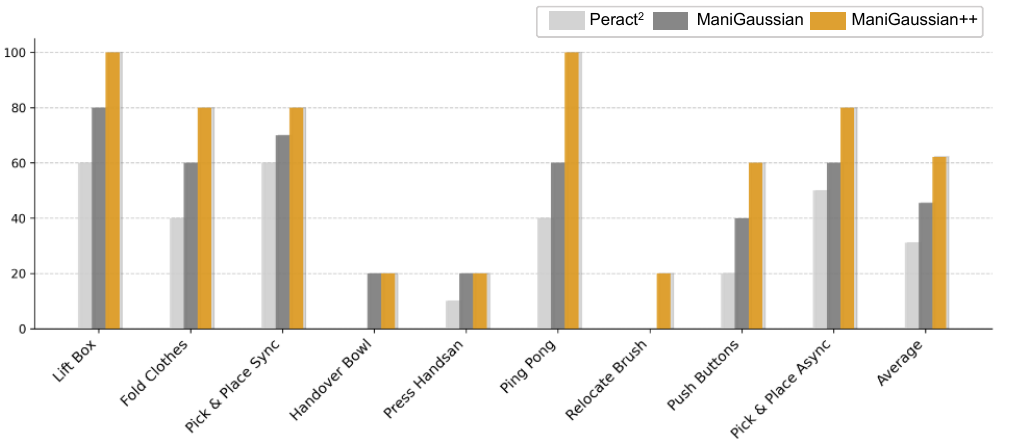}
    \caption{\small \textbf{Real-world Results.} We train and evaluate Peract$^2$, ManiGaussian and \method on $9$ challenging real-world tasks.}
    \vspace{-0.4cm}
    \label{fig:real_world_results}
\end{figure*}

\begin{figure}[t]
    \centering
    \includegraphics[width=0.47\textwidth]{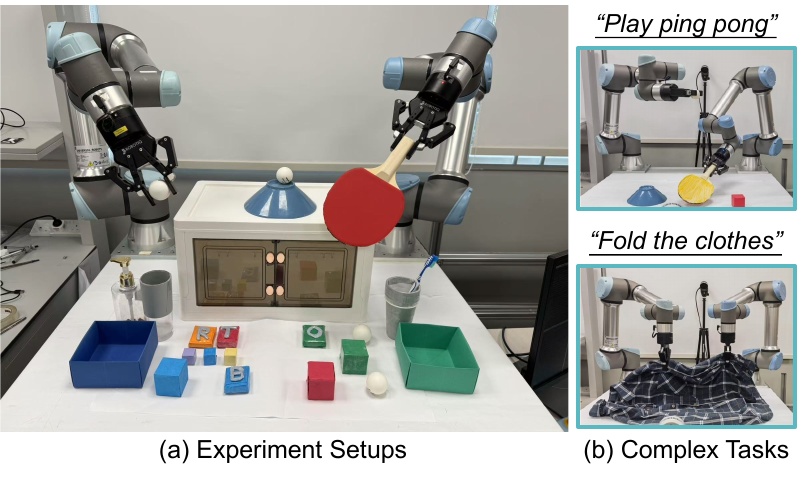}
    \caption{\textbf{Real-World Experiments} with two UR5e manipulators.
    }
    \vspace{-0.6cm}
    \label{fig:real_world_gallery}
\end{figure}

\subsection{Ablation Study}\label{subsec: ablation_study}
% The proposed concepts include task-oriented Gaussian Splatting that differentiates acting and stabilizing arms for multi-body spatiotemporal dynamics modeling, and hierarchical Gaussian world model that mines multi-body spatiotemporal dynamics for the visual representation. We verify the effectiveness of each proposed concept in \Cref{table:ablation_study}.
% We start with a vanilla PerAct$^2$ baseline with an average success rate of $5.67$\%, and then progressively include the proposed techniques.
% Specifically, introducing the Gaussian regressor to predict the Gaussian parameters improves the average performance by $14.33$\%. The task-oriented Gaussian Splatting advances the baseline by $21.00$\%, which verifies the significance of distinguishing the stablizing and acting roles for better arm collaboration. Finally, the design of hierarchical Gaussian world model boosts the performance by $33.33$\%, which demonstrates the effectiveness of digesting the multi-body dynamics for bimanual manipulation.

We propose task-oriented Gaussian Splatting to differentiate acting and stabilizing arms for multi-body spatiotemporal dynamics modeling, and hierarchical Gaussian world model for visual representation. \Cref{table:ablation_study} shows the effectiveness of each concept. We start with a vanilla PerAct$^2$ baseline ($5.67$\% success rate), and then progressively include the proposed techniques. Using a Gaussian regressor to predict parameters improves performance by $14.33$\% ($5.67$\% \text{vs} $20.00$\%). Task-oriented Gaussian Splatting advances the baseline by $21.00$\% ($5.67$\% \text{vs} $26.67$\%), emphasizing the importance of distinguishing arm roles for better collaboration. Finally, the hierarchical Gaussian world model boosts the performance by $33.33$\% ($26.67$\% \text{vs} $60.00$\%), which demonstrates the effectiveness of digesting the multi-body dynamics for bimanual manipulation.
\subsection{Qualitative Analysis}\label{subsec:qualitative}

% \noindent\textbf{Visualization of Whole Trajectories.}
% We present two qualitative examples of the generated action sequence in \Cref{fig:qualitative_case_study} from GNFactor and our ManiGaussian. 
% In this case, the agent is instructed to \emph{``Slide the block to yellow target''}. The results show that the previous agent struggles to complete the task since it imitates the expert's backward pulling motion, even though the claw is already leaning towards the right side of the red block.
% In contrast, \method returns to the red square and successfully slides the square to the yellow target, owing to that our method can correctly understand the scene dynamics of objects in contact. 

% \noindent\textbf{Visualization of Novel View Synthesis.}
\Cref{fig:qualitative_novel_view_synthesis} shows the novel view image synthesis results. First, based on the front view observation where the gripper shape cannot be seen, our \method offers superior detail in modeling buttons and grippers in novel views. Second, our method accurately predicts future states based on the recovered details. 
% For example, in the top case of the \texttt{press buttons} task, our \method not only predicts the future gripper position, but also predicts the future location of buttons influenced by the gripper. This qualitative result demonstrates that our \method learns the intricate multi-body dynamics successfully.
% In the bottom case, for the task \texttt{straighten rope}, \method is able to predict the proper coordination of both robot arms and the physical consequence of the rope that is straightened. These cases illustrates the effectiveness of the proposed hierarchical Gaussian world model.

For example, in the top case of the \texttt{press buttons} task, our \method is able to predict the current and the future gripper position. This qualitative result demonstrates that our \method learns the intricate multi-body dynamics successfully. In the bottom case, for the task \texttt{sweep to dustpan}, \method not only predicts the future position, but also predicts the future location of broom influenced by the gripper and the proper coordination of both robot arms. These cases illustrate the generation fidelity of the proposed hierarchical Gaussian world model.

\subsection{Real-world Experiments}\label{subsec:real_world_exp}
We validate Peract$^2$, ManiGaussian and \method with $9$ challenging real-robot tasks, which is depicted in \Cref{fig:real_world_results}, our \method outperform Peract$^2$ and ManiGaussian by a sizable relative improvement of  $100$\% ($31.11$\% \text{vs} $62.22$\%) and $36.57$\% ($45.56$\% \text{vs} $62.22$\%), shows that our method is able to complete all $9$ tasks simultaneously with only one model conditioned on natural human language from scratch, without any pertaining on the simulation or sim-to-real transferring.
Notably, \Cref{fig:real_world_gallery} shows that \method is able to complete complex tasks like \texttt{Play ping pong} and \texttt{Fold Clothes} that involve complex collaboration patterns, attributing to the proposed hierarchical Gaussian world model that encodes the multi-body spatiotemporal dynamics for the visual representation.
Besides, \method is robust to distractors like the lightning environment, which further validates the generalizability obtained by mining the multi-body spatiotemporal dynamics.
Please refer to supplementary videos for more real-world qualitative results and details on the task setups.

\section{Conclusion}
\label{sec:conclusion}

In this paper, we have presented \method, a novel framework that addresses the challenges of multi-task bimanual manipulation through hierarchical Gaussian world modeling. 
Our approach extends the ManiGaussian framework by explicitly modeling multi-body spatiotemporal dynamics via a hierarchical Gaussian world model for dual-arm collaboration. 
Specifically, 
% we generate task-oriented Gaussian Splatting from intermediate visual features to differentiate between acting and stabilizing arms for multi-body dynamics modeling. 
We use task-oriented Gaussian Splatting from visual features to differentiate acting and stabilizing arms for dynamics modeling.
% We then construct a hierarchical Gaussian world model with a leader-follower architecture, where the leader predicts Gaussian Splatting deformation caused by the stabilizing arm's motions, while the follower generates the physical consequences of the acting arm's movements. 
A hierarchical Gaussian world model employs a leader-follower architecture: the leader predicts deformation from the stabilizing arm, while the follower models the acting arm’s effects.
Through extensive experiments, \method demonstrates significant improvements over state-of-the-art general bimanual manipulation methods, achieving an improvement of 20.2\% in 10 simulated tasks and 60\% success rate in 9 challenging real-world tasks. 
Limitations include the demand for calibrated multi-view cameras for supervisions, which increases the cost of real robot deployment.
% \newpage
{
    % \small
    \bibliographystyle{IEEEtranS}
    \bibliography{main}
}

% WARNING: do not forget to delete the supplementary pages from your submission 
% \input{sec/X_suppl}

\end{document}